\documentclass{article}
\usepackage{arxiv}
\usepackage[utf8]{inputenc} 
\usepackage[T1]{fontenc}    
\usepackage{hyperref}       
\usepackage{url}            
\usepackage{booktabs}       
\usepackage{amsfonts}       
\usepackage{amssymb}        
\usepackage{nicefrac}       
\usepackage{microtype}      
\usepackage{lipsum}
\usepackage{multirow}
\usepackage{tabularx}
\usepackage{graphicx}
\usepackage[table]{xcolor}
\usepackage{tikz}
\usepackage{float}   
\usepackage{caption} 
\usetikzlibrary{positioning, shapes.geometric, arrows.meta, calc}
\definecolor{tier2}{RGB}{255,193,7}
\definecolor{tier3}{RGB}{220,53,69}
\graphicspath{ {./images/} }

\title{TrustX Agent Risk Classification Framework (ARC): Risk-Tiering Internally Created Agentic AI Systems}

\author{
 Hannah Liu \\
  Responsible AI Institute\\
  Imperial College London\\
  \texttt{hannah@responsible.ai} \\
   \And
 Rhea Saxena \\
  Responsible AI Institute\\
  \texttt{rhea@responsible.ai} \\
  \And
 Shiv Asthana \\
  Responsible AI Institute\\
  \texttt{shiv@responsible.ai} \\
}

\begin{document}
\maketitle
\begin{abstract}
The proliferation of agentic AI systems across enterprise and public-sector contexts has outpaced the capacity of general-purpose AI risk frameworks to classify and govern them. In this paper, we introduce the TrustX Agent Risk Classification Framework, a structured, repeatable instrument that can be applied to seven types of agentic AI systems and is grounded in foundational pre-existing AI governance frameworks. At the core of the framework is a twelve-dimension scoring rubric that robustly quantifies the risk. This rubric is combined with other components, such as the GPA + IAT classification model and the five-level autonomy framework derived from existing literature. These inputs produce a three-tier governance output with mapped control recommendations. A specialised Coding Assistant extension is also included to account for nuances specific to this type of agentic AI system. We then use an illustrative example to show our framework in practice. ARC is intended for AI governance practitioners, risk officers, developers, and regulators, and it will regularly undergo iteration as we continue to expand it and make it more robust. The community can access the interactive framework \href{https://arc.responsible.ai/}{here}.
\end{abstract}


\section{Introduction}
The deployment of agentic AI systems, or applications capable of advanced perception, planning, and action with relatively high degrees of autonomy and ability for orchestration \cite{Sapkota_2026,xi2023rise,wang2024survey}, has accelerated across enterprise operations, financial services, healthcare, software development, and public-sector administration. Recent surveys document the rapid expansion of LLM-based agents in research and industry \cite{xi2023rise,wang2024survey}, while empirical analyses caution that agent capabilities are proliferating faster than governance practices needed to evaluate them reliably \cite{kapoor2024agents}. 

This proliferation of agentic AI creates governance challenges that existing, general-purpose AI risk frameworks are not designed to address, a gap seen in early work on practices for governing agentic AI systems \cite{shavit2023practices}. For example, the NIST AI Risk Management Framework \cite{nistrm}, ISO/IEC 42001 \cite{iso42001}, OWASP guidelines \cite{owasp2025agenticthreats, owasp2026top10agentic}, and MITRE ATLAS techniques \cite{atlas} are influential, though their guidance is largely voluntary. Sector-specific instruments such as SR 11-7 and SR 26-2 help mandate model risk management in banking, though they have explicitly noted that their guidance does not cover generative and agentic AI \cite{sr117, sr262}. The closest to a binding regulatory document is the EU AI Act \cite{euaiact}, though its scope also fails to consider agentic AI due to it being published before the significant prevalence of agentic AI systems.

Concurrently, a series of real-world incidents have demonstrated the concrete risk of ungoverned agentic systems. In November 2025, Anthropic disclosed that a Chinese state-sponsored group known as GTG-1002 had manipulated Claude Code to conduct espionage operations against roughly 30 global targets, with the coding assistant agent executing around 80-90\% of the attack autonomously \cite{anthropic2025disrupting}. In December 2025, security researcher Ari Marzouk deemed his findings as the "IDEsaster" after discovering that there were over 30 vulnerabilities across more than ten AI coding tools, including remote code execution via JSON schema attacks and IDE settings manipulation \cite{marzouk2025idesaster, lakshmanan2025researchers}. In March 2025, Pillar Security disclosed the Rules File Backdoor attack, in which hidden Unicode characters were used in configuration files to manipulate Copilot and Cursor outputs \cite{karliner2025rulesfile}. 
Many such incidents instantiate a broader class of indirect prompt injection attacks, characterized in previous work such as Greshake et al. \cite{greshake2023not}, where adversarial instructions embedded in third-party content are executed by LLM-integrated applications.In 2026, we see this pattern of agentic AI incidents continue to occur, ranging from persistent prompt injection \cite{maloyan2026promptinjectionattacksagentic} to executing multi-stage attacks without significant human involvement \cite{lazer2026surveyagenticaicybersecurity}. 

In this paper, we aim to provide a practical recommendation for assessing the risks of agentic AI systems. We provide a complete intake and risk-tiering methodology, while also adopting operational definitions for agentic AI and agent autonomy from Bommarito \cite{bommarito2025} and Feng \cite{feng2025}. At the framework's core is a twelve-dimension risk scoring rubric that uses current governance frameworks \cite{euaiact, nistrm, iso42001, atlas, owasp2025agenticthreats, owasp2026top10agentic, sr117, sr262} as a grounded foundation. A three-tiered risk assessment output is calculated as a result of the weighting of all components of the framework, which uses a "critical dimension" approach that prevents high risks from being averaged out in the algorithm. The framework includes a specific Coding Assistant extension to account for further nuances found in this type of agentic AI system. 

\textbf{Our key contributions are as follows:}
\begin{enumerate}
    \item \textbf{Twelve-dimension risk scoring rubric} grounded in established frameworks, with a "critical dimension" approach that prevents high risks from being averaged away.
    \item \textbf{A specialised Coding Assistant extension} featuring a Capabilities Assessment that assesses properties specific to coding assistants, a Deployment Model Classification that accompanies the core autonomy level classification, and coding assistant-specific additional risk factors.
    \item \textbf{Comprehensive framework weighting formula} that effectively considers all assessed components to produce a risk-tiered output. 
\end{enumerate}

\section{Background and Related Work}
\subsection{Pre-Existing AI Risk Assessments and Governance Frameworks}
The ARC tool builds upon several foundational and established frameworks. The NIST AI RMF \cite{nistrm} provided the community with a structure that situates four functions at its core: Map, Measure, Manage, and Govern. The first three are localised components, whereas Govern is designed to be a function that influences and cross-cuts across the three other components. The Generative AI Profile (NIST AI 600-1) enumerates risks novel to or exacerbated by generative models \cite{nist2024genai}. The EU AI Act \cite{euaiact}, one of the most influential pieces of legislation in AI governance, establishes a four-tier risk classification with binding obligations for high risk systems. ISO/IEC 42001 \cite{iso42001} is the first international standard for AI management systems, providing applicable controls that individuals can use to address various issues in AI system deployment. OWASP has released frameworks regarding threat patterns that are specific to agentic AI architectures, such as the white paper on Agentic AI Threats and Mitigations \cite{owasp2025agenticthreats} and the Top 10 for Agentic Applications for 2026 \cite{owasp2026top10agentic}. MITRE ATLAS \cite{atlas} is a knowledge base that documents adversarial risks and techniques that target AI systems. Finally, the Federal Reserve guidance on model risk management through SR 11-7 and SR 26-2 \cite{sr117, sr262} provides specified guidance on AI-driven financial models. 

Beyond these established frameworks, there are several other developments in the field of AI governance and risk assessments. For example, the AURA Framework \cite{aura2025} addresses agent autonomy risk assessment by employing a modular approach and gamma-based risk scoring methodology. Another example is the Cloud Security Alliance's Capabilities-Based Risk Assessment (CBRA), which provides a scalable governance method that enterprises can utilise through its inclusion of four core risk dimensions \cite{Cloud_Security_Alliance_2025}. Work targeting agent governance specifically has emerged in the past couple years.
Chan et al. \cite{chan2024visibility} propose measures for agent visibility such as identifiers, real-time monitoring, activity logging, and later provide an outline for the infrastructure needed to mediate and attribute agent interactions \cite{chan2025infrastructure}. South et al. develop protocols for authenticated delegation, letting human principals guide an agent's authority \cite{south2025authenticated}, while Hammond et al. \cite{hammond2025multiagent} survey failure modes unique to multi-agent settings. Related efforts in taxonomy include the MIT AI Risk Repository \cite{slattery2024airisk}, Weidinger et al.'s \cite{weidinger2022taxonomy} taxonomy of language model risks and Raji et al.'s \cite{raji2020closing} framework for internal algorithmic auditing. Sandboxed and adversarial benchmarks like ToolEmu \cite{ruan2024toolemu}, AgentDojo \cite{debenedetti2024agentdojo} and AgentHarm \cite{andriushchenko2025agentharm} operationalise agent-risk measurement for evaluations. 

However, these existing frameworks lack several key components for agentic AI risk assessment. Established generic AI governance frameworks do not necessarily have the capacity to scope in agentic AI, whether it's due to their voluntary nature or when they were last published. More recent work engages agentic AI directly but tends to be either narrowly evaluative, as with the benchmarks above, or position-oriented, arguing for or against a course of action rather than offering a repeatable classification instrument. Mitchell et al. \cite{mitchell2025fully}, for example, argue that fully autonomous agents should not be developed at all.

\subsection{The Categorical Nuances of AI Coding Assistants}
AI coding assistants merit distinct treatment within any agentic risk framework. These tools descend from large language models trained on code, beginning with Codex \cite{chen2021evaluating}, and have evolved from autocomplete-style suggestion engines into agentic systems like SWE-agent \cite{yang2024sweagent}, which autonomously navigate repositories, edit files, and execute tests. Benchmarks such as SWE-bench show rapid gains on real-world software engineering tasks \cite{jimenez2024swebench}.
This evolution has a security cost. Pearce et al. \cite{pearce2022asleep} found that roughly 40\% of Copilot-generated programs in security-relevant scenarios contained vulnerabilities. Perry et al. \cite{perry2023users} showed in a controlled user study that developers with AI assistance wrote significantly less secure code while believing it to be more secure. Sandoval et al. \cite{sandoval2023lost} found similar results in a systems-programming user study. Hou et al. \cite{hou2025mcp} add that the Model Context Protocol, now the standard by which coding assistants access external tools, introduces its own security threats. Combined with the incidents described in Section 1 \cite{anthropic2025disrupting,marzouk2025idesaster,lakshmanan2025researchers,karliner2025rulesfile,maloyan2026promptinjectionattacksagentic}, this is why Section 4 introduces a dedicated Coding Assistant extension.
\subsection{Conceptual Foundations of Agentic AI}
The concept of agency in AI has evolved from early symbolic AI planners to today's foundation-model-powered agents capable of multi-step reasoning, tool use, and autonomous execution. 
Key technical milestones include ReAct \cite{yao2023react}, which interleaves reasoning traces with environment-changing actions; Toolformer \cite{schick2023toolformer}, which demonstrated that language models can teach themselves to invoke external tools; and multi-agent orchestration frameworks such as AutoGen \cite{wu2023autogen}, which compose multiple conversing agents into larger workflows. 
Bommarito et al. \cite{bommarito2025} capture this progression through the GPA + IAT properties model, which defines agency via six operational properties.

The first half of the properties, known as GPA, are the foundational requirements for Minimal Agency. Systems that lack these core properties are classified as Non-Agentic under Bommarito et al.'s model. G represents Goal, in which the agent pursues an objective. P represents Perception, in which the agent receives environmental information. A represents Action, in which the agent has the ability to affect the environment.

The second half of the properties, known as IAT, are what allow a system to constitute as Agentic Systems. I represents Iteration, in which the agent operates in a repeated perception-action cycle. A represents Adaptation, in which the agent has the ability to modify its behaviour based on user feedback. Finally, T represents Termination, in which the agent has a defined stopping condition. 

\subsection{Autonomy Levels in Agentic Systems}
Graduated levels of automation have a long history in human factors research, originating with Sheridan and Verplank's ten-level scale of automation for undersea teleoperation \cite{sheridan1978human} and later refined by Parasuraman et al. \cite{parasuraman2000model} into a model of types and levels of human interaction with automation. The SAE J3016 \cite{sae2021j3016} levels of driving automation demonstrate how such scales can anchor regulation and industry practice in a safety-critical domain, and Morris et al. \cite{morris2024levels} have recently proposed an analogous "Levels of AGI" ontology in which autonomy is a central axis of both capability and risk. 

Risk assessment requires us to consider the amount of freedom an agent has in its surroundings. To characterise the degree of autonomy an agent has in relation to its human user, we adopt Feng et al.'s \cite{feng2025} five-level Autonomy framework. The levels are as follows. L1: Operator refers to the user directing all of the agent's actions, while the agent acts on command. L2: Collaborator refers to the user and the agent collaboratively planning and executing actions. L3: Consultant refers to the agent leading the actions, but consulting the user for expertise and preferences. L4: Approver refers to the agent engaging the user only in situations where they require a green-light in specified risk scenarios. Finally, L5: Observer refers to the agent having full autonomy, whereas the user only has the ability to monitor. Higher autonomy levels demand correspondingly more rigorous governance controls.

\section{The Agent Risk Classification (ARC) Framework}
\subsection{Agentic AI Taxonomy}

Our taxonomy is informed by prior mappings of the agent design space, including Gabriel et al.'s \cite{gabriel2024ethics} analysis of the ethics of advanced AI assistants  and Chan et al.'s \cite{chan2023harms} account of harms from increasingly agentic algorithmic systems, which shows that both the likelihood and severity of harms scale with the degree of agency conferred on a system. 

In ARC, agent types are differentiated along four different axes: 1) primary function and task domain, 2) degree of autonomy and human-in/on-the-loop (HITL/HOTL) characteristics, 3) nature of the operating environment, and 4) its action authority and system reach. From these axes, we identified seven categories of agent types:
\begin{enumerate}
    \item \textbf{Autonomous agents} refer to self-directed, goal-pursuing systems with minimal human intervention. An example would be deep research agents.
    \item \textbf{Coding assistants} refer to agents we use to help us generate code, create a file structure for our codebase, or guide us in building our projects. Examples include Claude Code or Codex.
    \item \textbf{Decision support systems} refer to AI tools that provide users with information that aids with their decision-making. An example would be clinical decision-making tools that inform physicians on their next steps.
    \item \textbf{AI embedded/physical agents} operate in physical systems, such as vehicles, robots, and industrial equipment. An example would be an autonomous vehicle like Waymo.
    \item \textbf{Knowledge assistants} refer to AI tools that assist users with information retrieval and synthesis. An example would be a Q\&A agent. 
    \item \textbf{Tool-using agents} have the ability to invoke external APIs, tools, or services in their functionality. An example would be an agent using the Model Context Protocol to query databases.
    \item \textbf{Transaction/commerce agents} are capable of executing financial or commercial actions. Examples would be algorithmic trading agents or agentic commerce protocols. 
\end{enumerate}

Of these seven, coding assistants require additional structural extensions due to their position in the software supply chain, the executable nature of their outputs, and their ability for direct system access. As a result, it requires extensions from our core structure, which is discussed in Section 4.

\subsection{Framework Structure}
\subsubsection{Agent Identification}
The initial screen contains metadata fields that record the agent's system type, the agent's name, the agent's ID, the industry it is being used for, and the primary deployment region for the agent. This process allows the user to keep track of their assessed agents.

\subsubsection{GPA + IAT Agency Classification}
We employ the GPA + IAT framework \cite{bommarito2025} in our framework through a checklist assessment. The classification specified by Bommarito et al. is automated in the backend of the framework to capture the relationship between properties selected and agency level. While this classification does not directly contribute to the final risk tier calculation, it provides the user with an idea of the level of agency their tool has. This is crucial for the user's understanding, as highlighted by Bommarito et al. in their paper \cite{bommarito2025}. Lack of clarity in defining whether one's tool or system is truly agentic leads to several downstream failures in compliance and efficiency, motivating the addition of this framework in ARC. Table ~\ref{tab:gpaiat} summarises the six properties with supporting evidence. 

In several of the agent system types, a property may not be available to be clicked on. This is because that property may generally not be applicable to the system type. For example, Action may not be applicable to a Knowledge Assistant because these types of agents do not act on their environment. When a property is not applicable to an agent, the algorithm is adjusted to accurately capture agency regardless of the missing property. In this case, if Goal and Perception are fulfilled, the agent will still achieve Minimal Agency status. 

\begin{table}[!ht]
\centering
\caption{GPA+IAT properties assessment adapted from Bommarito et al. \cite{bommarito2025}}
\label{tab:gpaiat}
\footnotesize
\begin{tabularx}{\columnwidth}{@{}l l X@{}}
\toprule
\textbf{Tier} & \textbf{Property} & \textbf{Description} \\
\midrule
\multirow{3}{*}{\rotatebox{90}{\scriptsize GPA}}
& Goal & Agent pursues a defined objective \\
& Perception & Receives information from environment \\
& Action & Can affect its environment \\
\midrule
\multirow{3}{*}{\rotatebox{90}{\scriptsize IAT}}
& Iteration & Operates in repeated perceive-act cycles \\
& Adaptation & Modifies behaviour based on feedback \\
& Termination & Has defined stopping conditions \\
\bottomrule
\end{tabularx}
\end{table}

\subsubsection{Autonomy Level Assessment}
We employ Feng et al.'s five-level autonomy framework \cite{feng2025} in our assessment by allowing the user to select the most appropriate autonomy level. This is stored and taken in consideration during the final risk tier calculation. Levels 1 and 2 map to a low risk tier, levels 3 and 4 map to a medium risk tier, and level 5 maps to a high risk tier. 

\subsubsection{Twelve-Dimension Risk Scoring}
We developed twelve risk dimensions after synthesising the main insights derived from each of the foundational AI governance frameworks we selected. Each dimension is scored via a point scale of (1) Low, (2) Medium, and (3) High. Table~\ref{tab:twelve_dims} presents each of the twelve dimensions with tier descriptions. 

The framework then takes a sum of all risk dimension scores, before averaging them across the number of available risk dimensions. This formula ensures that if a risk dimension does not apply to an agent system type, it will not produce a skewed result. The highest individual dimension score is also stored, as this is important in our "critical dimension" approach when we calculate the final risk tier. 

\newcolumntype{Y}{>{\raggedright\arraybackslash}X}
\begin{table*}[t]
\centering
\caption{The twelve-dimension risk scoring rubric at the core of the ARC framework}
\label{tab:twelve_dims}
\footnotesize
\begin{tabularx}{\textwidth}{@{}>{\raggedright\arraybackslash}p{2cm} Y Y Y Y@{}}
\toprule
\textbf{Dimension} & \textbf{Tier 1 (Low)} & \textbf{Tier 2 (Medium)} & \textbf{Tier 3 (High)} & \textbf{Scoring Criteria} \\
\midrule
Autonomy & Human-in-the-loop required for every action & Human-on-the-loop; agent acts but human can intervene & Fully autonomous operation with no real-time oversight & Level of human supervision \\
\addlinespace
Decision Scope & Localised decisions for the specific task & Domain or function-level decisions & Enterprise-wide decisions & Influence on organisation \\
\addlinespace
Temporal Coupling & Isolated, independent actions & Chained workflows with potential domino effects & Continuous autonomous feedback loops & Risk compounding over time \\
\addlinespace
Action Authority & Read-only or advisory & Create/modify content & Full transaction execution authority & Scope of performable actions \\
\addlinespace
System Reach & Single internal system & Multiple internal systems & Cross-domain or third-party systems & Degree of system access \\
\addlinespace
Blast Radius & Single user or small scope & Team or department level & Enterprise-wide or public-facing & Scale of potential harm \\
\addlinespace
Persistence & Stateless; no memory & Session-based memory & Long-term persistent memory & State retention capabilities \\
\addlinespace
Reversibility & Fully reversible & Partially reversible & Irreversible or cascading & Ability to undo actions \\
\addlinespace
Control Authority & Standalone agent & Supervises other agents & Orchestrates agent fleets & Control over other agents \\
\addlinespace
Data Sensitivity & Public or non-sensitive & Internal or confidential & Regulated or crown-jewel data & Classification of accessed data \\
\addlinespace
Aggregation Risk & No aggregation & Limited aggregation & Cross-session or inferential & Harm through accumulation \\
\addlinespace
Data Egress Paths & Constrained outputs & Multiple controlled outputs & Multi-tool or external paths & Methods data can exit \\
\bottomrule
\end{tabularx}
\end{table*}

\subsubsection{Additional Risk Factors}
At the end of ARC framework, we include a checklist that has additional risk factors that do not necessarily fit into the twelve core risk dimensions, but are still valuable to keep in mind. While they do not contribute to the final risk tier calculation, they do contribute to the selection of policies and controls that ARC may recommend the user to consider and employ in their respective use cases. We consider the following additional risk factors: PII handling, financial transaction processing, credit/underwriting decisions, regulated data access, multi-agent coordination requirements, real-time decision-making, and external system integration.

\subsection{Final Risk Tier Determination Logic}
We use a "critical dimension" approach to ensure that a high-scoring risk dimension or risk component is not averaged away by our formulas. This allows us to employ the most risk-averse method in our framework, ensuring the highest level of proactive assessment. At the same time, averaging allows us to differentiate between the Medium and Low risk tiers, therefore supporting its inclusion. The approach is as follows:
\begin{itemize}
    \item \textbf{Tier 1: Low Risk.} This occurs when all dimensions = 1 and the average risk dimension score $< 1.5$.
    \item \textbf{Tier 2: Medium Risk.} This occurs when either the individual maximum dimension is 2, or if the average risk dimension score $\geq 1.5$. Autonomy levels 3 and 4 also lift the tier to medium risk if the scoring criteria is borderline between Medium and Low.
    \item \textbf{Tier 3: High Risk.} This occurs when any dimension scores a (3) High, or if the agent's autonomy level is L5.
\end{itemize}
Multi-agent systems inherit the highest tier among all of its agents, as their ability to orchestrate can easily propagate the effects of components from a high-risk agent to an otherwise low-risk one. Users should reassess their agentic system if there are any changes in its capabilities, tooling, or deployment context.

\subsection{Recommended Governance Controls}
Each risk tier has specific types of controls mapped to them. Controls are drawn from the ARC control catalog, which are created in alignment to our selection of established AI governance frameworks. The delineation is as follows:
\begin{itemize}
    \item \textbf{Tier 1: Standard.} Controls for this tier include examples such documentation, basic HITL/HOTL practices, and implementation of audit logs.
    \item \textbf{Tier 2: Enhanced.} Controls for this tier include examples such as model behaviour boundaries, kill switch implementation, and enhanced monitoring.
    \item \textbf{Tier 3: Rigorous.} Controls for this tier include examples such as third-party validation, continuous monitoring, and regulatory reporting. Agentic systems at this level may also require board-level approval before deployment.
\end{itemize}

\section{The ARC Coding Assistant Framework Extension}
\subsection{Capabilities Assessment}
Replacing the GPA + IAT component in the core ARC framework, we developed an assessment composed of 20 discrete capabilities, as seen in Table~\ref{tab:capabilities}. These were developed based on our research syntheses of coding assistants and their functions, while being stress-tested by members in our community. We provide this assessment in place of GPA + IAT to further detail out the coding assistant's profile with specificity to its unique properties. Despite these capabilities not being included in the final risk calculations, the coding assistant's capability profile directly informs the user when they eventually select choices for risk dimensions, autonomy level, and deployment model classification. 

\begin{table}[t]
\centering
\caption{Capabilities assessment for ARC coding assistant extension}
\label{tab:capabilities}
\scriptsize
\setlength{\tabcolsep}{4pt}
\begin{tabular}{@{}l l@{}}
\toprule
\textbf{Capability} & \textbf{Description} \\
\midrule
Code Generation & Generates new code from natural language prompts \\
Code Completion & Provides inline code suggestions and auto-completion \\
Code Review & Reviews code for quality, style, and best practices \\
Bug Detection & Identifies potential bugs and errors in code \\
Test Generation & Generates unit tests and test cases automatically \\
Documentation Generation & Creates documentation from code automatically \\
Refactoring Suggestions & Suggests code improvements and refactoring opportunities \\
Security Scanning & Scans code for security vulnerabilities \\
Performance Optimization & Identifies and suggests performance improvements \\
API Integration & Assists with API integration and consumption \\
Database Query Generation & Generates SQL and database queries \\
Infrastructure as Code & Generates IaC templates (Terraform, CloudFormation) \\
CI/CD Pipeline Assistance & Helps configure CI/CD pipelines \\
Version Control Integration & Integrates with Git and version control systems \\
Multi-language Support & Supports multiple programming languages \\
Framework-specific Assistance & Provides framework-specific guidance and code \\
Code Translation & Translates code between programming languages \\
Debugging Assistance & Helps debug and troubleshoot code issues \\
Dependency Management & Manages package dependencies and versions \\
Architecture Recommendations & Provides architecture and design pattern suggestions \\
\bottomrule
\end{tabular}
\end{table}

\subsection{Deployment Model Classification}
In addition to the L1-L5 autonomy level component in the core ARC framework, we have added four deployment models that help users to further define and assess the autonomy of their coding assistant (Table ~\ref{tab:deployment}). Although deployment model classification does not feed into the final risk calculation, it helps user verify their autonomy level of choice by grounding it in coding assistant examples. This allows for users to be better informed in the risk assessment process by providing checks before submitting and receiving the final result. 

\begin{table}[!ht]
\centering
\caption{Coding assistant deployment models}
\label{tab:deployment}
\footnotesize
\begin{tabularx}{\columnwidth}{@{}l X c@{}}
\toprule
\textbf{Model} & \textbf{Description} & \textbf{Expected Tier Mapping} \\
\midrule
1: IDE Autocomplete & Suggestions only; no autonomous execution or file modifications & Tier 1 \\
\addlinespace
2: File-Level Agent & Can modify files, run read commands; requires approval & Tier 2 \\
\addlinespace
3: Autonomous Multi-Step & Executes 30+ minute workflows; may skip approvals & Tier 2--3 \\
\addlinespace
4: Production-Connected & Access to production, cloud, or regulated systems & Tier 3 \\
\bottomrule
\end{tabularx}
\end{table}

\subsection{Coding Assistant-Specific Risk Factors}
The original risk dimensions in the core ARC are preserved in the coding assistant extension. However, the extension replaces the generic additional risk factors of the core framework with a coding assistant-specific risk taxonomy of 20 factors organised into three categories (Table~\ref{tab:codingrisks}). The categories follow the path a coding assistant's outputs and actions travel. It starts at what the assistant receives and produces, then moves to how its behaviour can be redirected, and finally ending and what it is permitted to have access to. We deliberately grounded the taxonomy in the disclosed incidents that we discussed in the Introduction and Background. 

During the assessment, the user selects every factor applicable to the coding assistant, while also being guided by the earlier capabilities and deployment model classification sections. For example, a Deployment Model 1 with its suggestion-only context is less likely to exhibit command injection via terminal than a Deployment Model 4. 

\begin{table}[t]
\centering
\caption{Coding assistant-specific risk factors}
\label{tab:codingrisks}
\scriptsize
\setlength{\tabcolsep}{4pt}
\begin{tabular}{@{}l l@{}}
\toprule
\textbf{Category} & \textbf{Risk factor} \\
\midrule
\multirow{5}{*}{\begin{tabular}[c]{@{}l@{}}Supply Chain and\\Artifact Integrity\end{tabular}}
  & Supply chain attack vector \\
  & Training data poisoning \\
  & Package hallucination/squatting \\
  & License compliance issues \\
  & Insecure code generation (CWEs) \\
\midrule
\multirow{10}{*}{\begin{tabular}[c]{@{}l@{}}Agent Manipulation and\\Workflow Hijacking\end{tabular}}
  & IDE configuration manipulation \\
  & Invisible Unicode character attack \\
  & Rules file backdoor \\
  & Persona manipulation attack \\
  & Multi-agent coordination attack \\
  & Multi-root workspace exploitation \\
  & Data exfiltration via generated code \\
  & JSON schema remote trigger \\
  & Autonomous workflow without bounds \\
  & Prompt injection vulnerability \\
\midrule
\multirow{5}{*}{\begin{tabular}[c]{@{}l@{}}Access and\\Privilege Abuse\end{tabular}}
  & Production system access \\
  & Permission bypass (``YOLO'') modes \\
  & Code review bypass \\
  & Secrets/credentials leakage \\
  & Command injection via terminal \\
\bottomrule
\end{tabular}
\end{table}

\section{Illustrative Examples with Agent-Type Risk Profiles}
To demonstrate the ARC framework in practice, we apply it to representative configurations and profiles of each agent type. The scores we yielded are illustrative and reflect typical deployments rather than specific assessed systems, as we recognise that user developed systems of each type may contain nuances that our general assessment may not cover. As a result, we do not go into additional risk factors, as those are highly specific to each use case. Ultimately, this section is intended to demonstrate how the methodology produces differentiated risk tiers across agent types. Table~\ref{tab:crossagent} shows the result of our illustrative exercise.

\subsection{Autonomous Agents}
Autonomous agents are self-directed and pursue user-set goals with minimal human intervention. Examples of autonomous agents that we used in our hypothetical scoring included operations optimisation agents that monitor deployed systems, customer service agents that independently resolve tickets, and strategic planning agents that execute multi-step business processes. 

Applying them to the GPA +IAT assessment showed that these types of systems consistently yield the Agentic System classification status, as all six properties are regularly present in these applications. For autonomy level, the majority of these tools would identify with L5 because they handle full autonomous operation procedures. 

Risk scoring using the twelve-dimension rubric reveals a total score of 28 and an average of 2.33. The highest risk dimensions are Autonomy, Temporal Coupling, Persistence, and Data Sensitivity, reflecting the ability of these agents to continuously monitor and act on deployed systems that operate in autonomous feedback loops, leading to potential risk compounding over time. 

We would situate autonomous agents generally at Tier 3: High Risk. This majorly stems from their high autonomy level and them scoring a 3 on various risk dimensions. The elevated average score also confirms that risk is found broadly across the rubric rather than concentrated. ARC would assign rigorous controls to autonomous agents in this general scenario, including aspects such as third-party validation, continuous monitoring, and regulatory reporting.

\subsection{Decision Support Systems}
Decision support systems generate insights and recommendations for users to aid in their decision-making. An example of what we used in our hypothetical scoring is a Q\&A agent.

The GPA + IAT assessment yields an Agentic System classification in most scenarios, though this mainly depends on whether Adaptation and Iteration are present for a particular decision support system. When it comes to autonomy level, we chose to classify it low, such as around L1 to L2. This is because a defining property of a decision support system is that a human makes the final decision, whereas the agent acts on user command and produces outputs that are mostly advisory in nature. 

Applying it to the risk dimensions yielded a total score of 17 and an average of 1.42. Although this places it below the 1.5 average threshold that separates the low and medium tiers, our profile places Data Sensitivity at a 3, since some types of decision support systems (such as clinical or financial) operate over regulated data. Despite several other dimensions scoring only a 1, the single high-scoring dimension places this agent system type at Tier 3: High Risk. This example demonstrates our "critical dimension" approach, as a simple averaging formula would mask the concentrated exposure that regulated data access represents. 

\subsection{AI Embedded/Physical Agents}
Physical agents include autonomous vehicles, surgical robots, and industrial automation. Despite their innovative nature, they present unique risks due to the irreversibility of their actions and their high blast radius in safety-critical environments.

In the GPA + IAT assessment, all six properties are present generally in physical agents, making them Agentic Systems. Some physical agents, such as autonomous vehicles, are capable of executing actions and interacting with the environment with only a human monitoring it. As a result, in the riskiest case scenario, physical agents would have an autonomy level of L5. 

On the risk dimension rubric, the total score is 23, and the average is 1.92. Autonomy, Action Authority, Blast Radius, and Reversibility all scored a 3, which reflects how physical agents operate in public-facing, safety-critical environments and execute actions that cannot necessarily be undone. As a result, we would place it at Tier 3: High Risk.

\subsection{Knowledge Assistants}
Knowledge assistants, such as customer-facing FAQ agents or enterprise RAG chatbots over internal documentation, are built to help retrieve and synthesise information for the user. 

On the GPA + IAT classification, knowledge assistants can range from Minimal Agency to Agentic System depending on Adaptation capabilities. Some knowledge assistants do not take user feedback into consideration, whereas some do. The autonomy level of a knowledge assistant is typically lower, such as around L1 to L2. This is because in most cases, a user is directing commands that the agent responds to.

On the risk dimensions, there was a total score of 14 and an average of 1.17. All other dimensions, except for Persistence and Data Sensitivity, scored a 1. This reflects how these kinds of agents do not necessarily take any actions, does not reach many external systems, and rarely makes decisions. However, they still have session-based memory and may have access to internal documentation. These observations have knowledge assistants ranking at Tier 2: Medium Risk.

However, a common knowledge assistant variant worth acknowledging is a RAG assistant with cross-session personalisation operating in setting with regulated data. In this case, Persistence and Data Sensitivity would elevate it to Tier 3: High Risk. This distinction highlights the need for the audience to perform ARC on specific systems rather than assuming them from agent system type. 

\subsection{Tool-Using Agents}
Tool-using agents, such as ones that use the Model Context Protocol to complete various functions, invoke external APIs and services in its tasks. 

The GPA + IAT assessment places it as fulfilling all six properties, meaning that generally, tool-using agents are Agentic Systems. However, the autonomy level is where there is more ambiguity. While some agents only propose tool calls and require human approval to move forward, others are capable of fully autonomous tool orchestration. For our hypothetical scenario, we will choose a profile that sits somewhere in the middle, such as a tool-using agent with a HOTL configuration. This would place its autonomy level at L3. 

The risk dimensions, however, show the tendency for a tool-using agent to be a high risk system in general. There was a total score of 25 with an average of 2.08, with System Reach and Data Egress Paths scoring a 3. This is because a defining property of tool-using agents is their ability to invoke third-party APIs and services, which also allows for several external paths through which data can unintentionally leak or exit. Driven by these two dimensions, this agent system type would rank at Tier 3: High Risk. 

\subsection{Transaction/Commerce Agents}
Transaction and commerce agents are typically used by enterprises to execute purchasing, payments, claims, approval, and inventory management. 

On the GPA + IAT assessment, transaction and commerce agents in general have all six properties, making them Agentic Systems. For autonomy levels, the majority of these kinds of agents would have a level of L3 because the agents tend to rely on human preferences while still executing actions autonomously.

Risk scoring yields a total of 15 and an average of 1.25, with no dimensions scoring at a 3. The highest dimensions are Autonomy,  Persistence, and Data Sensitivity at a 2, which reflects how these agents can use a retained transaction and account history across sessions. As a result, this agent system type would rank at Tier 2: Medium Risk. 

When we consider this in conjunction with the regulatory landscape surrounding AI financial services use cases, we recognise that Tier 2: Medium Risk is arguably the lowest risk level for any customer-facing transaction agent. This is due to model risk management guidance under SR 11-7 and SR 26-2 \cite{sr117, sr262} and general consumer protection guidelines. Any additional extensions would easily escalate the transaction/commerce agent's risk profile, an event that is common for enterprise-scope agentic commerce deployments. 

\subsection{Coding Assistants}
Coding assistants assist users in creating codebases and building projects, therefore having a range of capabilities that are normally determined by its user. Because of its unique properties, they follow an extension of ARC. In accordance to our deployment model component, we will use an IDE autocomplete tool (M1) as our illustrative profile.

For the Capabilities Assessment, an IDE autocomplete tool would exhibit code completion, code generation, multi-language support, framework-specific assistance, documentation generation, test generation, and refactoring suggestions. Every selected capability reflects the advisory delivery mechanism of an IDE autocomplete tool, while capabilities focused on analysis, execution, and integration are absent.

With the system begin classified as Model 1, it would most likely be mapped to an autonomy level of L1. This is because a developer usually directs the activities of an IDE autocomplete tool and is the one approving and enacting every change. The tool's suggestions carry no effect unless they are accepted.

On the risk dimensions, it produced a total of 14 and an average of 1.17. The highest scoring dimensions are Persistence and Data Sensitivity, which were both at a 2. This reflects session-based context over an open codebase and the possibility of exposing the IDE autocomplete tool to internal, confidential source code. Given that the maximum dimension is 2, we would rank the IDE autocomplete tool at Tier 2: Medium Risk.

\begin{table*}[t]
\centering
\caption{Illustrative risk score comparison by agent system type}
\label{tab:crossagent}
\scriptsize
\setlength{\tabcolsep}{3pt}
\resizebox{\textwidth}{!}{%
\begin{tabular}{@{}l c *{12}{c} r r c@{}}
\toprule
\textbf{Agent Type}
  & \rotatebox{75}{\textbf{Auton.\ Level}}
  & \rotatebox{75}{\textbf{Auton.}}
  & \rotatebox{75}{\textbf{Dec.\ Scope}}
  & \rotatebox{75}{\textbf{Temporal}}
  & \rotatebox{75}{\textbf{Action}}
  & \rotatebox{75}{\textbf{Sys.\ Reach}}
  & \rotatebox{75}{\textbf{Blast Rad.}}
  & \rotatebox{75}{\textbf{Persist.}}
  & \rotatebox{75}{\textbf{Revers.}}
  & \rotatebox{75}{\textbf{Control}}
  & \rotatebox{75}{\textbf{Data Sens.}}
  & \rotatebox{75}{\textbf{Aggreg.}}
  & \rotatebox{75}{\textbf{Egress}}
  & \textbf{Total}
  & \textbf{Avg}
  & \textbf{Tier} \\
\midrule
Autonomous          & L5 & \cellcolor{tier3!30}3 & 2 & \cellcolor{tier3!30}3 & 2 & 2 & 2 & \cellcolor{tier3!30}3 & 2 & 2 & \cellcolor{tier3!30}3 & 2 & 2 & 28 & 2.33 & \cellcolor{tier3!30}\textbf{3} \\
Coding Asst.\ (M1)  & L1 & 1 & 1 & 1 & 1 & 1 & 1 & 2 & 1 & 1 & 2 & 1 & 1 & 14 & 1.17 & \cellcolor{tier2!30}\textbf{2} \\
Decision Support    & L1--L2 & 1 & 2 & 1 & 1 & 1 & 2 & 1 & 1 & 1 & \cellcolor{tier3!30}3 & 2 & 1 & 17 & 1.42 & \cellcolor{tier3!30}\textbf{3} \\
Embedded/Physical   & L5 & \cellcolor{tier3!30}3 & 1 & 2 & \cellcolor{tier3!30}3 & 1 & \cellcolor{tier3!30}3 & 2 & \cellcolor{tier3!30}3 & 1 & 2 & 1 & 1 & 23 & 1.92 & \cellcolor{tier3!30}\textbf{3} \\
Knowledge Asst.     & L1--L2 & 1 & 1 & 1 & 1 & 1 & 1 & 2 & 1 & 1 & 2 & 1 & 1 & 14 & 1.17 & \cellcolor{tier2!30}\textbf{2} \\
Tool-Using          & L3 & 2 & 2 & 2 & 2 & \cellcolor{tier3!30}3 & 2 & 2 & 2 & 1 & 2 & 2 & \cellcolor{tier3!30}3 & 25 & 2.08 & \cellcolor{tier3!30}\textbf{3} \\
Transaction         & L3 & 2 & 1 & 1 & 1 & 1 & 1 & 2 & 1 & 1 & 2 & 1 & 1 & 15 & 1.25 & \cellcolor{tier2!30}\textbf{2} \\
\bottomrule
\end{tabular}%
}
\end{table*}

\section{Discussion}
\subsection{Implications}
The ARC framework aims to be a practical stepping stone in creating proactive and robust risk assessments for agentic AI systems. With many of the current governance frameworks providing the academic community with foundations that are useful yet mostly voluntary and performative, we offer an interactive framework that the community can use to directly check whether their applications are compliant with established guidelines and AI governance structures. Instead of having to estimate whether one's agentic tools and systems fit into the obligations and requirements of a governance framework, we provide a way for users to do so concretely. 

We hope that organisations will take aspects of the ARC framework and embed it into procurement gates, development life-cycle reviews, and deployment approvals. For example, organisations can make it a part of their evaluation process when key reassessment triggers occur, such as if the tool has new integrations, deployment model transitions, or data access scope. 

\subsection{Is ARC Too Restrictive?}
One argument the ARC framework may receive is that the "critical dimension" approach may lead it to be too restrictive, as any dimension that scores a 3 automatically lifts an agentic system to the highest risk tier. This high level of governance may discourage innovative tools, which commonly include high-risk but high-reward aspects that can significantly forward an organisation or industry's capabilities. An example would be ambient voice technology in the NHS, which was introduced despite its high-risk features \cite{NHS_England_2025}. Today, it is a key piece of software that drastically improves clinical efficiency.

On the other hand, we argue that increased governance and stricter approaches will end up encouraging better innovation in the longer time horizon. In particular, our "critical dimension" approach offers a design principle that is transferable to policy. We believe that no amount of low-risk properties should offset a single high-risk property because it does not provide adequate awareness towards the issues that truly matter. If we flag high-risk properties of innovative products early in their life-cycle, developers and enterprise leaders may be able to create guardrails that help the longevity of the product. It may also encourage people to create in accordance to governance, because in the long-term, systems and tools that do not adhere to AI governance guidelines and policies will not survive when that legislation becomes official and more binding. As a result, we crafted ARC to follow these insights. However, with AI governance being a growing field, we welcome any and all thoughts on this topic.

\subsection{Limitations}
One primary limitation of ARC is scoring subjectivity, which we aimed to mitigate as much as possible through detailed rubrics and guidance. However, we acknowledge that differences in user interpretation can lead to varying results, and we welcome community feedback on improving definitions and examples that we provide in our \href{https://arc.responsible.ai/}{interactive framework website.} Another limitation is that the framework provides a static snapshot and does not necessarily capture drift over the agent life-cycle. To mitigate this, we provide users with reassessment triggers throughout the paper and generally recommend running the framework again when any significant changes occur to the agentic system. Finally, with the rapid speed of the agentic AI landscape, the framework may become outdated quickly as more agentic configurations and abilities emerge. 

\subsection{Future Work}
We plan to expand this framework to include multi-agent systems, a new yet increasingly prevalent concept within agentic AI. We also plan to include different risk surfaces, such as procurement and external exposure risk, in future iterations of the framework to ensure further robustness of the risk assessment's capabilities. Tailoring our framework to specific industries is also part of our future research plans, as our framework currently works best with financial services and healthcare use cases. Finally, we intend to embed an iterative and versioning mechanism into the framework to ensure its guidelines, risk dimensions, and other components stay up to date with new developments in the AI governance space. 

\section{Conclusion}
The Agent Risk Classification framework provides a structured and repeatable methodology for risk-tiering agentic AI systems. Its components allow the framework to address the full spectrum of agentic AI systems currently being deployed in enterprise contexts, while also accounting for nuances in specific types of agentic AI systems. In its embedded formulas, our "critical dimension" approach ensures that high-risk properties are recognised and not diluted through averaging. Our outputs then feed directly into informing policies and controls, therefore situating ARC as an integrated instrument that bridges the gap between theoretical AI risk frameworks and the operational requirements of policymakers, regulators, and governance practitioners. We invite adoption, feedback, and contribution to future versions from the AI governance community.

\bibliographystyle{unsrt}  
\bibliography{references} 

@article{bommarito2025,
  title={What is an Agent? A Conceptual Primer and History of Agents and Agentic AI},
  author={Bommarito, Michael James and Bommarito, Jillian and Katz, Daniel Martin},
  journal={A Conceptual Primer and History of Agents and Agentic AI (November 25, 2025)},
  year={2025},
  url = {https://dx.doi.org/10.2139/ssrn.5806982}
}

@misc{feng2025,
  title={Levels of Autonomy for AI Agents}, 
      author={K. J. Kevin Feng and David W. McDonald and Amy X. Zhang},
      year={2025},
      eprint={2506.12469},
      archivePrefix={arXiv},
      primaryClass={cs.HC},
      url={https://arxiv.org/abs/2506.12469},
}

@misc{nistrm,
  author = {{National Institute of Standards and Technology}},
  title = {AI Risk Management Framework (AI RMF 1.0)},
  year = {2023},
  url = {https://www.nist.gov/itl/ai-risk-management-framework}
}

@misc{euaiact,
  author = {{European Union}},
  title = {Regulation (EU) 2024/1689 on Artificial Intelligence},
  year = {2024},
  url = {https://eur-lex.europa.eu/eli/reg/2024/1689/oj/eng}
}

@misc{iso42001,
  author = {{ISO/IEC}},
  title = {ISO/IEC 42001:2023 Artificial Intelligence Management System Standard},
  year = {2023},
  url = {https://www.iso.org/standard/42001}
}

@techreport{owasp2025agenticthreats,
  author       = {{OWASP Gen AI Security Project}},
  title        = {Agentic {AI} -- Threats and Mitigations},
  institution  = {OWASP Foundation},
  year         = {2025},
  month        = {2},
  version      = {1.0},
  url          = {https://genai.owasp.org/download/45674/?tmstv=1739819891},
  urldate      = {2026-07-05},
  note         = {First guide in the OWASP Agentic Security Initiative (ASI) series}
}

@techreport{owasp2026top10agentic,
  author       = {{OWASP Gen AI Security Project}},
  title        = {{OWASP} Top 10 for Agentic Applications 2026},
  institution  = {OWASP Foundation},
  year         = {2025},
  month        = {12},
  url          = {https://genai.owasp.org/download/52117/?tmstv=1765059207},
  urldate      = {2026-07-05},
  note         = {Published December 9, 2025}
}

@misc{atlas,
  author = {{MITRE}},
  title = {MITRE ATLAS: Adversarial Threat Landscape for AI Systems},
  year = {2024},
  url = {https://atlas.mitre.org/techniques}
}

@misc{sr117,
  author = {{Board of Governors of the Federal Reserve System}},
  title = {SR 11-7: Guidance on Model Risk Management},
  year = {2011},
  url = {https://www.federalreserve.gov/supervisionreg/srletters/sr1107.htm}
}

@misc{sr262,
  author = {{Board of Governors of the Federal Reserve System and Office of the Comptroller of the Currency and Federal Deposit Insurance Corporation}},
  title = {SR 26-2: Revised Guidance on Model Risk Management},
  year = {2026},
  url = {https://www.federalreserve.gov/supervisionreg/srletters/SR2602.htm}
}

@misc{aura2025,
  title={AURA: An Agent Autonomy Risk Assessment Framework}, 
      author={Lorenzo Satta Chiris and Ayush Mishra},
      year={2025},
      eprint={2510.15739},
      archivePrefix={arXiv},
      primaryClass={cs.AI},
      url={https://arxiv.org/abs/2510.15739},
}

@article{Sapkota_2026,
   title = {AI Agents vs. Agentic AI: A Conceptual taxonomy, applications and challenges},
journal = {Information Fusion},
volume = {126},
pages = {103599},
year = {2026},
issn = {1566-2535},
doi = {https://doi.org/10.1016/j.inffus.2025.103599},
url = {https://www.sciencedirect.com/science/article/pii/S1566253525006712},
author = {Ranjan Sapkota and Konstantinos I. Roumeliotis and Manoj Karkee}
}

@techreport{anthropic2025disrupting,
  author       = {{Anthropic}},
  title        = {Disrupting the First Reported AI-Orchestrated Cyber Espionage Campaign},
  institution  = {Anthropic},
  year         = {2025},
  month        = {11},
  url          = {https://www-cdn.anthropic.com/d7dd50dd1185f59be051b307150d877f2b82bd2c.pdf},
}

@misc{marzouk2025idesaster,
  author       = {Marzouk, Ari},
  title        = {{IDEsaster}: A Novel Vulnerability Class in {AI} {IDEs}},
  url = {https://maccarita.com/posts/idesaster/},
  year         = {2025},
  month        = {12},
}

@misc{lakshmanan2025researchers,
  author       = {Lakshmanan, Ravie},
  title        = {Researcher Uncovers 30+ Flaws in {AI} Coding Tools Enabling Data Theft and {RCE} Attacks},
  year         = {2025},
  month        = {12},
  url          = {https://thehackernews.com/2025/12/researchers-uncover-30-flaws-in-ai.html},
}

@misc{karliner2025rulesfile,
  author       = {Karliner, Ziv},
  title        = {New Vulnerability in {GitHub} {Copilot} and {Cursor}: How Hackers Can Weaponize Code Agents},
  year         = {2025},
  month        = {3},
  url          = {https://www.pillar.security/blog/new-vulnerability-in-github-copilot-and-cursor-how-hackers-can-weaponize-code-agents},
}

@misc{maloyan2026promptinjectionattacksagentic,
      title={Prompt Injection Attacks on Agentic Coding Assistants: A Systematic Analysis of Vulnerabilities in Skills, Tools, and Protocol Ecosystems}, 
      author={Narek Maloyan and Dmitry Namiot},
      year={2026},
      eprint={2601.17548},
      archivePrefix={arXiv},
      primaryClass={cs.CR},
      url={https://arxiv.org/abs/2601.17548}, 
}

@misc{lazer2026surveyagenticaicybersecurity,
      title={A Survey of Agentic AI and Cybersecurity: Challenges, Opportunities and Use-case Prototypes}, 
      author={Sahaya Jestus Lazer and Kshitiz Aryal and Maanak Gupta and Elisa Bertino},
      year={2026},
      eprint={2601.05293},
      archivePrefix={arXiv},
      primaryClass={cs.CR},
      url={https://arxiv.org/abs/2601.05293}, 
}

@misc{Cloud_Security_Alliance_2025, 
    title={Capabilities-based risk assessment (CBRA) for AI Systems}, 
    url={https://cloudsecurityalliance.org/artifacts/capabilities-based-risk-assessment-cbra-for-ai-systems}, 
    journal={Cloud Security Alliance}, 
    author={Cloud Security Alliance}, 
    year={2025}, 
    month={Nov},
}

@techreport{shavit2023practices,
  author      = {Shavit, Yonadav and Agarwal, Sandhini and Brundage, Miles and Adler, Steven and O'Keefe, Cullen and Campbell, Rosie and Lee, Teddy and Mishkin, Pamela and Eloundou, Tyna and Hickey, Alan and Slama, Katarina and Ahmad, Lama and McMillan, Paul and Vallone, Alex and Passos, Alexandre and Robinson, David G.},
  title       = {Practices for Governing Agentic {AI} Systems},
  institution = {OpenAI},
  year        = {2023},
  month       = {12},
  url         = {https://openai.com/index/practices-for-governing-agentic-ai-systems/}
}

@misc{xi2023rise,
  title={The Rise and Potential of Large Language Model Based Agents: A Survey}, 
      author={Zhiheng Xi and Wenxiang Chen and Xin Guo and Wei He and Yiwen Ding and Boyang Hong and Ming Zhang and Junzhe Wang and Senjie Jin and Enyu Zhou and Rui Zheng and Xiaoran Fan and Xiao Wang and Limao Xiong and Yuhao Zhou and Weiran Wang and Changhao Jiang and Yicheng Zou and Xiangyang Liu and Zhangyue Yin and Shihan Dou and Rongxiang Weng and Wensen Cheng and Qi Zhang and Wenjuan Qin and Yongyan Zheng and Xipeng Qiu and Xuanjing Huang and Tao Gui},
      year={2023},
      eprint={2309.07864},
      archivePrefix={arXiv},
      primaryClass={cs.AI},
      url={https://arxiv.org/abs/2309.07864}, 
}

@article{wang2024survey,
  title={A survey on large language model based autonomous agents},
   volume={18},
   ISSN={2095-2236},
   url={http://dx.doi.org/10.1007/s11704-024-40231-1},
   DOI={10.1007/s11704-024-40231-1},
   number={6},
   journal={Frontiers of Computer Science},
   publisher={Springer Science and Business Media LLC},
   author={Wang, Lei and Ma, Chen and Feng, Xueyang and Zhang, Zeyu and Yang, Hao and Zhang, Jingsen and Chen, Zhiyuan and Tang, Jiakai and Chen, Xu and Lin, Yankai and Zhao, Wayne Xin and Wei, Zhewei and Wen, Jirong},
   year={2024},
   month=Mar 
}

@misc{kapoor2024agents,
  title={AI Agents That Matter}, 
      author={Sayash Kapoor and Benedikt Stroebl and Zachary S. Siegel and Nitya Nadgir and Arvind Narayanan},
      year={2024},
      eprint={2407.01502},
      archivePrefix={arXiv},
      primaryClass={cs.LG},
      url={https://arxiv.org/abs/2407.01502},
}

@inproceedings{greshake2023not,
  author = {Greshake, Kai and Abdelnabi, Sahar and Mishra, Shailesh and Endres, Christoph and Holz, Thorsten and Fritz, Mario},
title = {Not What You've Signed Up For: Compromising Real-World LLM-Integrated Applications with Indirect Prompt Injection},
year = {2023},
isbn = {9798400702600},
publisher = {Association for Computing Machinery},
address = {New York, NY, USA},
url = {https://doi.org/10.1145/3605764.3623985},
doi = {10.1145/3605764.3623985},
booktitle = {Proceedings of the 16th ACM Workshop on Artificial Intelligence and Security},
pages = {79–90},
numpages = {12},
location = {Copenhagen, Denmark},
series = {AISec '23}
}

@misc{nist2024genai,
  author = {Chloe Autio and Reva Schwartz and Jesse Dunietz and Shomik Jain and Martin Stanley and Elham Tabassi and Patrick Hall and Kamie Roberts},
  title = {Artificial Intelligence Risk Management Framework: Generative Artificial Intelligence Profile},
  year = {2024},
  month = {2024-07-26 04:07:00},
  publisher = {NIST Trustworthy and Responsible AI, National Institute of Standards and Technology, Gaithersburg, MD},
  url = {https://tsapps.nist.gov/publication/get_pdf.cfm?pub_id=958388},
  doi = {https://doi.org/10.6028/NIST.AI.600-1},
  language = {en},
}

@inproceedings{chan2024visibility,
  author = {Chan, Alan and Ezell, Carson and Kaufmann, Max and Wei, Kevin and Hammond, Lewis and Bradley, Herbie and Bluemke, Emma and Rajkumar, Nitarshan and Krueger, David and Kolt, Noam and Heim, Lennart and Anderljung, Markus},
title = {Visibility into AI Agents},
year = {2024},
isbn = {9798400704505},
publisher = {Association for Computing Machinery},
address = {New York, NY, USA},
url = {https://doi.org/10.1145/3630106.3658948},
doi = {10.1145/3630106.3658948},
booktitle = {Proceedings of the 2024 ACM Conference on Fairness, Accountability, and Transparency},
pages = {958–973},
numpages = {16},
location = {Rio de Janeiro, Brazil},
series = {FAccT '24}
}

@misc{chan2025infrastructure,
  title={Infrastructure for AI Agents}, 
      author={Alan Chan and Kevin Wei and Sihao Huang and Nitarshan Rajkumar and Elija Perrier and Seth Lazar and Gillian K. Hadfield and Markus Anderljung},
      year={2025},
      eprint={2501.10114},
      archivePrefix={arXiv},
      primaryClass={cs.AI},
      url={https://arxiv.org/abs/2501.10114}
}

@misc{south2025authenticated,
  title={Authenticated Delegation and Authorized AI Agents}, 
      author={Tobin South and Samuele Marro and Thomas Hardjono and Robert Mahari and Cedric Deslandes Whitney and Dazza Greenwood and Alan Chan and Alex Pentland},
      year={2025},
      eprint={2501.09674},
      archivePrefix={arXiv},
      primaryClass={cs.CY},
      url={https://arxiv.org/abs/2501.09674},
}

@misc{hammond2025multiagent,
  title={Multi-Agent Risks from Advanced AI}, 
      author={Lewis Hammond and Alan Chan and Jesse Clifton and Jason Hoelscher-Obermaier and Akbir Khan and Euan McLean and Chandler Smith and Wolfram Barfuss and Jakob Foerster and Tomáš Gavenčiak and The Anh Han and Edward Hughes and Vojtěch Kovařík and Jan Kulveit and Joel Z. Leibo and Caspar Oesterheld and Christian Schroeder de Witt and Nisarg Shah and Michael Wellman and Paolo Bova and Theodor Cimpeanu and Carson Ezell and Quentin Feuillade-Montixi and Matija Franklin and Esben Kran and Igor Krawczuk and Max Lamparth and Niklas Lauffer and Alexander Meinke and Sumeet Motwani and Anka Reuel and Vincent Conitzer and Michael Dennis and Iason Gabriel and Adam Gleave and Gillian Hadfield and Nika Haghtalab and Atoosa Kasirzadeh and Sébastien Krier and Kate Larson and Joel Lehman and David C. Parkes and Georgios Piliouras and Iyad Rahwan},
      year={2025},
      eprint={2502.14143},
      archivePrefix={arXiv},
      primaryClass={cs.MA},
      url={https://arxiv.org/abs/2502.14143},
}

@article{slattery2024airisk,
  title={The AI risk repository: A meta-review, database, and taxonomy of risks from artificial intelligence},
   volume={7},
   ISSN={2666-3899},
   url={http://dx.doi.org/10.1016/j.patter.2026.101517},
   DOI={10.1016/j.patter.2026.101517},
   number={5},
   journal={Patterns},
   publisher={Elsevier BV},
   author={Slattery, Peter and Saeri, Alexander K. and Grundy, Emily A.C. and Graham, Jess and Noetel, Michael and Uuk, Risto and Dao, James and Pour, Soroush and Casper, Stephen and Thompson, Neil},
   year={2026},
   month=May, pages={101517}
}

@inproceedings{weidinger2022taxonomy,
  author = {Weidinger, Laura and Uesato, Jonathan and Rauh, Maribeth and Griffin, Conor and Huang, Po-Sen and Mellor, John and Glaese, Amelia and Cheng, Myra and Balle, Borja and Kasirzadeh, Atoosa and Biles, Courtney and Brown, Sasha and Kenton, Zac and Hawkins, Will and Stepleton, Tom and Birhane, Abeba and Hendricks, Lisa Anne and Rimell, Laura and Isaac, William and Haas, Julia and Legassick, Sean and Irving, Geoffrey and Gabriel, Iason},
title = {Taxonomy of Risks posed by Language Models},
year = {2022},
isbn = {9781450393522},
publisher = {Association for Computing Machinery},
address = {New York, NY, USA},
url = {https://doi.org/10.1145/3531146.3533088},
doi = {10.1145/3531146.3533088},
booktitle = {Proceedings of the 2022 ACM Conference on Fairness, Accountability, and Transparency},
pages = {214–229},
numpages = {16},
location = {Seoul, Republic of Korea},
series = {FAccT '22}
}

@inproceedings{raji2020closing,
  author = {Raji, Inioluwa Deborah and Smart, Andrew and White, Rebecca N. and Mitchell, Margaret and Gebru, Timnit and Hutchinson, Ben and Smith-Loud, Jamila and Theron, Daniel and Barnes, Parker},
title = {Closing the AI accountability gap: defining an end-to-end framework for internal algorithmic auditing},
year = {2020},
isbn = {9781450369367},
publisher = {Association for Computing Machinery},
address = {New York, NY, USA},
url = {https://doi.org/10.1145/3351095.3372873},
doi = {10.1145/3351095.3372873},
booktitle = {Proceedings of the 2020 Conference on Fairness, Accountability, and Transparency},
pages = {33–44},
numpages = {12},
location = {Barcelona, Spain},
series = {FAT* '20}
}

@misc{mitchell2025fully,
  title={Fully Autonomous AI Agents Should Not be Developed}, 
      author={Margaret Mitchell and Avijit Ghosh and Alexandra Sasha Luccioni and Giada Pistilli},
      year={2025},
      eprint={2502.02649},
      archivePrefix={arXiv},
      primaryClass={cs.AI},
      url={https://arxiv.org/abs/2502.02649},
}

@misc{ruan2024toolemu,
  title={Identifying the Risks of LM Agents with an LM-Emulated Sandbox}, 
      author={Yangjun Ruan and Honghua Dong and Andrew Wang and Silviu Pitis and Yongchao Zhou and Jimmy Ba and Yann Dubois and Chris J. Maddison and Tatsunori Hashimoto},
      year={2024},
      eprint={2309.15817},
      archivePrefix={arXiv},
      primaryClass={cs.AI},
      url={https://arxiv.org/abs/2309.15817}
}

@misc{debenedetti2024agentdojo,
  title={AgentDojo: A Dynamic Environment to Evaluate Prompt Injection Attacks and Defenses for LLM Agents}, 
      author={Edoardo Debenedetti and Jie Zhang and Mislav Balunović and Luca Beurer-Kellner and Marc Fischer and Florian Tramèr},
      year={2024},
      eprint={2406.13352},
      archivePrefix={arXiv},
      primaryClass={cs.CR},
      url={https://arxiv.org/abs/2406.13352},
}

@misc{andriushchenko2025agentharm,
  title={AgentHarm: A Benchmark for Measuring Harmfulness of LLM Agents}, 
      author={Maksym Andriushchenko and Alexandra Souly and Mateusz Dziemian and Derek Duenas and Maxwell Lin and Justin Wang and Dan Hendrycks and Andy Zou and Zico Kolter and Matt Fredrikson and Eric Winsor and Jerome Wynne and Yarin Gal and Xander Davies},
      year={2025},
      eprint={2410.09024},
      archivePrefix={arXiv},
      primaryClass={cs.LG},
      url={https://arxiv.org/abs/2410.09024},
}

@misc{chen2021evaluating,
  title={Evaluating Large Language Models Trained on Code}, 
      author={Mark Chen and Jerry Tworek and Heewoo Jun and Qiming Yuan and Henrique Ponde de Oliveira Pinto and Jared Kaplan and Harri Edwards and Yuri Burda and Nicholas Joseph and Greg Brockman and Alex Ray and Raul Puri and Gretchen Krueger and Michael Petrov and Heidy Khlaaf and Girish Sastry and Pamela Mishkin and Brooke Chan and Scott Gray and Nick Ryder and Mikhail Pavlov and Alethea Power and Lukasz Kaiser and Mohammad Bavarian and Clemens Winter and Philippe Tillet and Felipe Petroski Such and Dave Cummings and Matthias Plappert and Fotios Chantzis and Elizabeth Barnes and Ariel Herbert-Voss and William Hebgen Guss and Alex Nichol and Alex Paino and Nikolas Tezak and Jie Tang and Igor Babuschkin and Suchir Balaji and Shantanu Jain and William Saunders and Christopher Hesse and Andrew N. Carr and Jan Leike and Josh Achiam and Vedant Misra and Evan Morikawa and Alec Radford and Matthew Knight and Miles Brundage and Mira Murati and Katie Mayer and Peter Welinder and Bob McGrew and Dario Amodei and Sam McCandlish and Ilya Sutskever and Wojciech Zaremba},
      year={2021},
      eprint={2107.03374},
      archivePrefix={arXiv},
      primaryClass={cs.LG},
      url={https://arxiv.org/abs/2107.03374},
}

@inproceedings{pearce2022asleep,
  author={Pearce, Hammond and Ahmad, Baleegh and Tan, Benjamin and Dolan-Gavitt, Brendan and Karri, Ramesh},
  booktitle={2022 IEEE Symposium on Security and Privacy (SP)}, 
  title={Asleep at the Keyboard? Assessing the Security of GitHub Copilot’s Code Contributions}, 
  year={2022},
  volume={},
  number={},
  pages={754-768},
  doi={10.1109/SP46214.2022.9833571}
}

@inproceedings{perry2023users,
  author = {Perry, Neil and Srivastava, Megha and Kumar, Deepak and Boneh, Dan},
title = {Do Users Write More Insecure Code with AI Assistants?},
year = {2023},
isbn = {9798400700507},
publisher = {Association for Computing Machinery},
address = {New York, NY, USA},
url = {https://doi.org/10.1145/3576915.3623157},
doi = {10.1145/3576915.3623157},
booktitle = {Proceedings of the 2023 ACM SIGSAC Conference on Computer and Communications Security},
pages = {2785–2799},
numpages = {15},
location = {Copenhagen, Denmark},
series = {CCS '23}
}

@misc{sandoval2023lost,
  title={Lost at C: A User Study on the Security Implications of Large Language Model Code Assistants}, 
      author={Gustavo Sandoval and Hammond Pearce and Teo Nys and Ramesh Karri and Siddharth Garg and Brendan Dolan-Gavitt},
      year={2023},
      eprint={2208.09727},
      archivePrefix={arXiv},
      primaryClass={cs.CR},
      url={https://arxiv.org/abs/2208.09727},
}

@misc{jimenez2024swebench,
  title={SWE-bench: Can Language Models Resolve Real-World GitHub Issues?}, 
      author={Carlos E. Jimenez and John Yang and Alexander Wettig and Shunyu Yao and Kexin Pei and Ofir Press and Karthik Narasimhan},
      year={2024},
      eprint={2310.06770},
      archivePrefix={arXiv},
      primaryClass={cs.CL},
      url={https://arxiv.org/abs/2310.06770},
}

@misc{yang2024sweagent,
  title={SWE-agent: Agent-Computer Interfaces Enable Automated Software Engineering}, 
      author={John Yang and Carlos E. Jimenez and Alexander Wettig and Kilian Lieret and Shunyu Yao and Karthik Narasimhan and Ofir Press},
      year={2024},
      eprint={2405.15793},
      archivePrefix={arXiv},
      primaryClass={cs.SE},
      url={https://arxiv.org/abs/2405.15793},
}

@misc{hou2025mcp,
  title={Model Context Protocol (MCP): Landscape, Security Threats, and Future Research Directions}, 
      author={Xinyi Hou and Yanjie Zhao and Shenao Wang and Haoyu Wang},
      year={2025},
      eprint={2503.23278},
      archivePrefix={arXiv},
      primaryClass={cs.CR},
      url={https://arxiv.org/abs/2503.23278},
}

@misc{yao2023react,
  title={ReAct: Synergizing Reasoning and Acting in Language Models}, 
      author={Shunyu Yao and Jeffrey Zhao and Dian Yu and Nan Du and Izhak Shafran and Karthik Narasimhan and Yuan Cao},
      year={2023},
      eprint={2210.03629},
      archivePrefix={arXiv},
      primaryClass={cs.CL},
      url={https://arxiv.org/abs/2210.03629},
}

@misc{schick2023toolformer,
  title={Toolformer: Language Models Can Teach Themselves to Use Tools}, 
      author={Timo Schick and Jane Dwivedi-Yu and Roberto Dessì and Roberta Raileanu and Maria Lomeli and Luke Zettlemoyer and Nicola Cancedda and Thomas Scialom},
      year={2023},
      eprint={2302.04761},
      archivePrefix={arXiv},
      primaryClass={cs.CL},
      url={https://arxiv.org/abs/2302.04761},
}

@misc{wu2023autogen,
  title={AutoGen: Enabling Next-Gen LLM Applications via Multi-Agent Conversation}, 
      author={Qingyun Wu and Gagan Bansal and Jieyu Zhang and Yiran Wu and Beibin Li and Erkang Zhu and Li Jiang and Xiaoyun Zhang and Shaokun Zhang and Jiale Liu and Ahmed Hassan Awadallah and Ryen W White and Doug Burger and Chi Wang},
      year={2023},
      eprint={2308.08155},
      archivePrefix={arXiv},
      primaryClass={cs.AI},
      url={https://arxiv.org/abs/2308.08155},
}

@inproceedings{sheridan1978human,
  title={Human and Computer Control of Undersea Teleoperators},
  author={Thomas B. Sheridan and William L. Verplank},
  year={1978},
  url={https://api.semanticscholar.org/CorpusID:106615927}
}

@article{parasuraman2000model,
  author={Parasuraman, R. and Sheridan, T.B. and Wickens, C.D.},
  journal={IEEE Transactions on Systems, Man, and Cybernetics - Part A: Systems and Humans}, 
  title={A model for types and levels of human interaction with automation}, 
  year={2000},
  volume={30},
  number={3},
  pages={286-297},
  doi={10.1109/3468.844354}
}

@techreport{sae2021j3016,
  author      = {{SAE International}},
  title       = {Taxonomy and Definitions for Terms Related to Driving Automation Systems for On-Road Motor Vehicles ({SAE} {J3016}\_202104)},
  institution = {SAE International},
  year        = {2021},
  url         = {https://www.sae.org/standards/content/j3016_202104/}
}

@misc{morris2024levels,
  title={Levels of AGI for Operationalizing Progress on the Path to AGI}, 
      author={Meredith Ringel Morris and Jascha Sohl-Dickstein and Noah Fiedel and Tris Warkentin and Allan Dafoe and Aleksandra Faust and Clement Farabet and Shane Legg},
      year={2025},
      eprint={2311.02462},
      archivePrefix={arXiv},
      primaryClass={cs.AI},
      url={https://arxiv.org/abs/2311.02462},
}

@misc{gabriel2024ethics,
  title={The Ethics of Advanced AI Assistants}, 
      author={Iason Gabriel and Arianna Manzini and Geoff Keeling and Lisa Anne Hendricks and Verena Rieser and Hasan Iqbal and Nenad Tomašev and Ira Ktena and Zachary Kenton and Mikel Rodriguez and Seliem El-Sayed and Sasha Brown and Canfer Akbulut and Andrew Trask and Edward Hughes and A. Stevie Bergman and Renee Shelby and Nahema Marchal and Conor Griffin and Juan Mateos-Garcia and Laura Weidinger and Winnie Street and Benjamin Lange and Alex Ingerman and Alison Lentz and Reed Enger and Andrew Barakat and Victoria Krakovna and John Oliver Siy and Zeb Kurth-Nelson and Amanda McCroskery and Vijay Bolina and Harry Law and Murray Shanahan and Lize Alberts and Borja Balle and Sarah de Haas and Yetunde Ibitoye and Allan Dafoe and Beth Goldberg and Sébastien Krier and Alexander Reese and Sims Witherspoon and Will Hawkins and Maribeth Rauh and Don Wallace and Matija Franklin and Josh A. Goldstein and Joel Lehman and Michael Klenk and Shannon Vallor and Courtney Biles and Meredith Ringel Morris and Helen King and Blaise Agüera y Arcas and William Isaac and James Manyika},
      year={2024},
      eprint={2404.16244},
      archivePrefix={arXiv},
      primaryClass={cs.CY},
      url={https://arxiv.org/abs/2404.16244},
}

@inproceedings{chan2023harms,
  author = {Chan, Alan and Salganik, Rebecca and Markelius, Alva and Pang, Chris and Rajkumar, Nitarshan and Krasheninnikov, Dmitrii and Langosco, Lauro and He, Zhonghao and Duan, Yawen and Carroll, Micah and Lin, Michelle and Mayhew, Alex and Collins, Katherine and Molamohammadi, Maryam and Burden, John and Zhao, Wanru and Rismani, Shalaleh and Voudouris, Konstantinos and Bhatt, Umang and Weller, Adrian and Krueger, David and Maharaj, Tegan},
title = {Harms from Increasingly Agentic Algorithmic Systems},
year = {2023},
isbn = {9798400701924},
publisher = {Association for Computing Machinery},
address = {New York, NY, USA},
url = {https://doi.org/10.1145/3593013.3594033},
doi = {10.1145/3593013.3594033},
booktitle = {Proceedings of the 2023 ACM Conference on Fairness, Accountability, and Transparency},
pages = {651–666},
numpages = {16},
location = {Chicago, IL, USA},
series = {FAccT '23}
}

@misc{NHS_England_2025, 
title={AI-enabled ambient scribing products in health and care settings},
url={https://www.england.nhs.uk/long-read/ai-enabled-ambient-scribing-products-in-health-and-care-settings/}, 
journal={NHS England}, 
author={NHS England}, 
year={2025}, 
month={Apr}}

\clearpage
\appendix
\section{ARC Framework Flow}
\begin{center}
\resizebox{!}{0.72\textheight}{%
\begin{tikzpicture}[
  font=\small,
  node distance=7mm and 10mm,
  box/.style={rectangle, rounded corners=2pt, draw=black!70, thick,
              fill=blue!8, text width=34mm, minimum height=9mm, align=center},
  cbox/.style={box, fill=orange!12},
  decision/.style={diamond, aspect=2.2, draw=black!70, thick, fill=black!4,
                   text width=22mm, align=center, inner sep=1pt},
  tierbox/.style={rectangle, rounded corners=2pt, draw=black!70, thick,
                  text width=22mm, minimum height=9mm, align=center},
  arrow/.style={-{Stealth[length=2.5mm]}, thick, black!70}
]
 
\node[box, fill=black!6] (intake) {\textbf{Identification}\\ \scriptsize system intake and scoping};
\node[decision, below=of intake] (branch) {\scriptsize Coding assistant?};
 
\node[box, below left=9mm and 14mm of branch] (gpa) {\textbf{GPA + IAT classification}\\ \scriptsize Non-Agentic / Minimal Agency / Agentic System};
\node[box, below=of gpa] (auton) {\textbf{Autonomy level}\\ \scriptsize L1--L5};
\node[box, below=of auton] (dims) {\textbf{Twelve risk dimensions}\\ \scriptsize scored 1--3};
\node[box, below=of dims] (factors) {\textbf{Additional risk factors}};
 
\node[cbox, below right=9mm and 14mm of branch] (cap) {\textbf{Capabilities assessment}\\ \scriptsize 20 discrete capabilities};
\node[cbox, below=of cap] (cauton) {\textbf{Autonomy level}\\ \scriptsize L1--L5};
\node[cbox, below=of cauton] (cdims) {\textbf{Twelve risk dimensions}\\ \scriptsize scored 1--3};
\node[cbox, below=of cdims] (deploy) {\textbf{Deployment model}\\ \scriptsize Models 1--4};
\node[cbox, below=of deploy] (crisk) {\textbf{Coding risk taxonomy}\\ \scriptsize 20 factors, 3 categories};
 
\node[box, fill=black!6] (tiercalc) at ($(branch |- crisk.south)+(0,-16mm)$)
  {\textbf{Tier determination}\\ \scriptsize weighting formula with critical-dimension rule};
 
\node[tierbox, fill=green!15, below left=8mm and 16mm of tiercalc] (t1) {\textbf{Tier 1}\\ \scriptsize Low Risk};
\node[tierbox, fill=tier2!25, below=8mm of tiercalc] (t2) {\textbf{Tier 2}\\ \scriptsize Medium Risk};
\node[tierbox, fill=tier3!25, below right=8mm and 16mm of tiercalc] (t3) {\textbf{Tier 3}\\ \scriptsize High Risk};
 
\node[box, fill=black!6, below=26mm of tiercalc] (controls)
  {\textbf{Tier-mapped controls}\\ \scriptsize monitoring, validation, and reporting requirements};
 
\draw[arrow] (intake) -- (branch);
\draw[arrow] (branch) -| node[pos=0.22, above, font=\scriptsize]{no} (gpa);
\draw[arrow] (branch) -| node[pos=0.22, above, font=\scriptsize]{yes} (cap);
 
\draw[arrow] (gpa) -- (auton);
\draw[arrow] (auton) -- (dims);
\draw[arrow] (dims) -- (factors);
 
\draw[arrow] (cap) -- (cauton);
\draw[arrow] (cauton) -- (cdims);
\draw[arrow] (cdims) -- (deploy);
\draw[arrow] (deploy) -- (crisk);
 
\draw[arrow] (factors.south) |- ($(tiercalc.north)+(0,4mm)$) -- (tiercalc.north);
\draw[arrow] (crisk.south) |- ($(tiercalc.north)+(0,4mm)$) -- (tiercalc.north);
 
\draw[arrow] (tiercalc) -| (t1);
\draw[arrow] (tiercalc) -- (t2);
\draw[arrow] (tiercalc) -| (t3);
 
\draw[arrow] (t1) |- (controls);
\draw[arrow] (t2) -- (controls);
\draw[arrow] (t3) |- (controls);
 
\end{tikzpicture}%
}

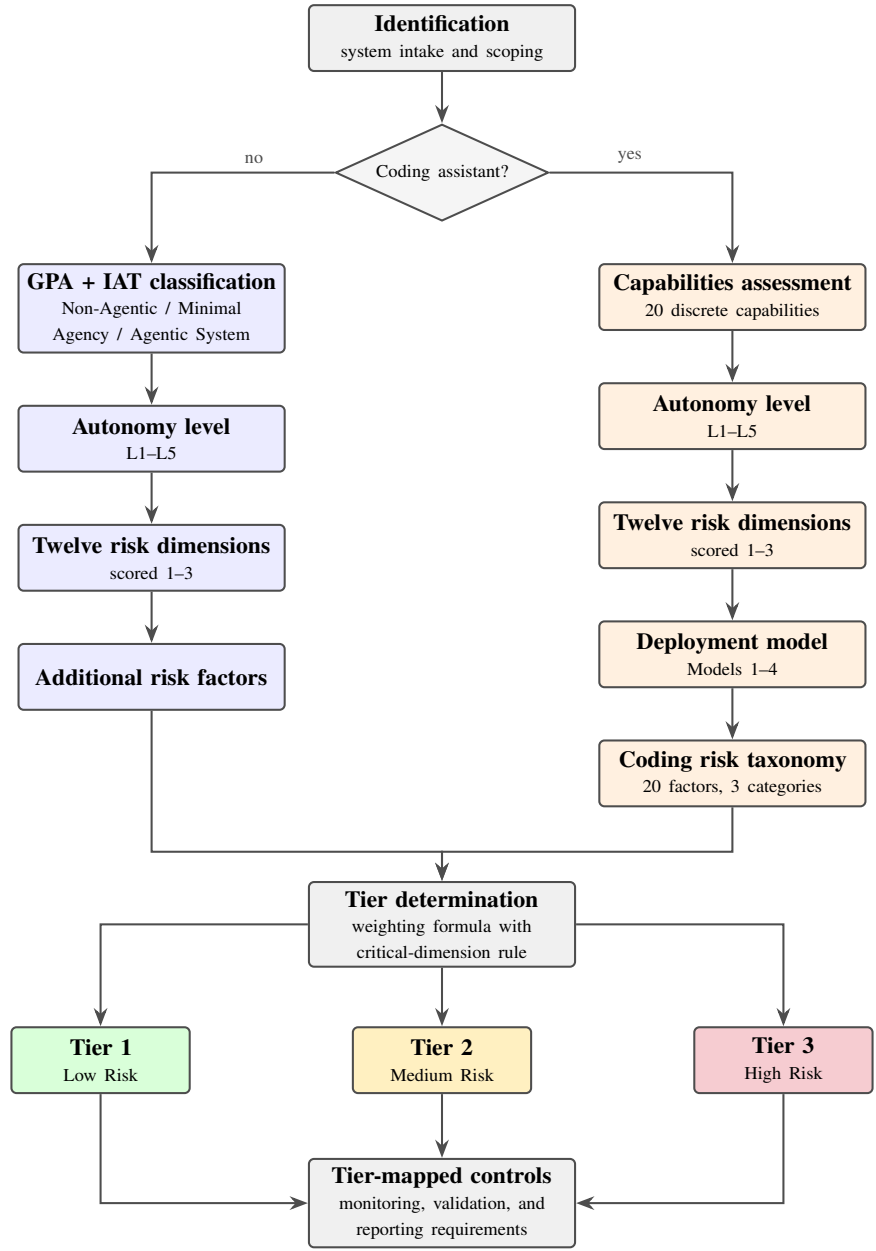
\captionof{figure}{The ARC assessment flow. All systems complete identification and intake; coding assistants are routed through the extension (right), which substitutes the capabilities assessment for GPA + IAT classification and adds deployment model classification and the coding-specific risk taxonomy. Both pathways converge on tier determination, where the weighting formula and critical-dimension rule produce a three-tier output mapped to controls.}
\label{fig:arcflow}
\end{center}

\end{document}